\documentclass[11pt]{article}
\usepackage{colacl06}
\usepackage{times}
\usepackage{latexsym}
\usepackage{fancybox}

\title{Goal-oriented Dialog as a Collaborative Subordinated Activity\\involving Collective Acceptance}

\author{Sylvie Saget\\
  IRISA - Projet CORDIAL \\
  6, rue de K\'erampont - BP 80518 \\
  22305 Lannion \\
  {\tt Sylvie.Saget@enssat.fr}  \And
  Marc Guyomard\\
  IRISA - Projet CORDIAL \\
  6, rue de K\'erampont - BP 80518 \\
  22305 Lannion \\
  {\tt Marc.Guyomard@enssat.fr}}

\date{}

\begin{document}
\maketitle
\begin{abstract}
Modeling dialog as a collaborative activity consists notably in specifying the content of the Conversational Common Ground and the kind of
social mental state involved. In previous work \cite{Sag06}, we claim that Collective Acceptance is the proper social attitude
for modeling Conversational Common Ground in the particular case of goal-oriented dialog. We provide a formalization of Collective
Acceptance, besides elements in order to integrate this attitude in a rational model of dialog are provided; and finally, a model of
referential acts as being part of a collaborative activity is provided. The particular case of reference has been chosen in order to
exemplify our claims.
\end{abstract}

\section{Introduction}

\indent Considering dialog as a collaborative activity is commonly admitted \cite{Cla96,GP04,CL91,CL94}.
Generally speaking, modeling a particular collaborative activity requires the specification of the collective intention helds
by the agents concerned and requires the specification of the Common Ground linked to this activity. Common Ground refers to pertinent
knowledge, beliefs and assumptions that are shared among team members \cite{Cla96}. Thus, Common Ground is a collection of social mental attitudes.
\newline
\newline
\indent The Common Ground linked to the dialogue itself (the Conversational Common Ground, CCG) ensures the mutual understanding
of dialog partners. The CCG enables dialog partners to use abbreviated forms of communication and enables them to be confident that
potentially
ambiguous messages will be correctly understood \cite{KFW05}. Dialogue partners become aligned at several linguistics
aspects \cite{GP04}. There is an alignment, for example, of the situation model, of the lexical and the syntactic levels,
even of clarity of articulation, of accent and of speech rate. Interactive alignment, of team members' situation model and of
social representations, facilitates language processing during conversation and facilitates social interaction.
\newline
\indent In the particular case of referent treatment, even for daily task, which use well-known objects with common known
proper names to refer to, there is a wide range of possible manners to describe this object by words. To ensure mutual understanding,
humans \emph{"associate objects with expressions (and the perspectives they encode), or else from achieving conceptual pacts, or
temporary, flexible agreements to view an object in a particular way"} \cite{BC96}.
\newline
\indent Thus, the Conversational Common Ground, since dialog is a mediated activity, contains all grounded elements linked to the way to
communicate (as the necessary level of clarity of articulation or speech rate) as well as elements of dialog's history such as
association between modes of presentation (linguistic objects) and mental representations: associations as conceptual pacts.
\newline
\newline
\indent In previous work \cite{Sag06}, we claim that Collective Acceptance is the proper social attitude for modeling Conversational Common
Ground in the particular case of goal-oriented dialog. In the first part of this paper, we show that such a modelization fits better than
stronger mental attitudes (such as shared beliefs or weaker epistemic states based on nested beliefs). We also show that
this modelization may be considered as partly due to the subordinated nature of goal-oriented dialog. Then, in the last part of the paper, a formalization of Collective Acceptance and elements are given in order
to integrate this attitude in a rational model of dialog. Finally a model of referential acts as being part of a collaborative activity is
provided. The particular case of reference has been chosen in order to exemplify our claims.

\section{Collective Acceptance: the proper social attitude for modeling CCG}

    \subsection{General claims on reference}

\indent In order to model dialog as a collaboration, reference resolution has to be considered as the \emph{"act identifying what the speaker
intends to be picked out by a noun phrase"} \cite{CL94}. Moreover, the collaborative nature of reference have been brought to the forefront
\cite{CWG86}. More precisely, reference is not the simple sum of the individual acts of generating and understanding, but is a collaborative
activity involving dialog partners. Thus, according to H.H. Clark et al. in \cite{CB04}, these individual acts are motivated by two
interrelated goals:  \label{GoalOfRefer}
\begin{itemize}
    \item[$\bullet$] Identification: Speakers are trying to get their addressees to identify a particular referent under a
    particular description.
    \item[$\bullet$] Grounding: Speakers and their addresses are trying to establish that the addressees have identified
    the referent as well enough for current purpose.
\end{itemize}

\indent How the identification goal is achieved ? First at all, when speaker has the intention to refer to a particular object, he has to
choose a description of this object. Traditionally, this choice is viewed as depending on the beliefs of dialog participants and as
depending on
availability. In other words, speaker can refer with a definite description $\imath x. \phi(x)$ to an object $o$ iff it is in the unique
available object for which $\phi(o)$ holds. Moreover, H.H. Clark and C.R. Marshall \cite{CM81} claimed that mutual knowledge of $\phi(o)$ is necessary,
if a description should refer successfully to an object $o$.\newline
\newline
\indent For example, let's imagine that two persons, Tom and Laura, who have been to the same school. Tom suggests to
Laura: "Shall we meet in front of our ex-school's basketball court". The choice of the description of the intented place
should be explained by the fact that Tom thinks that the following mutual belief is part of their common ground:

\begin{itemize}
    \item[$\bullet$] $MBel_{Tom, Laura} ( frontOf(l,h)$\newline
                        $ \wedge$ $basketballCourt(h)$\newline
                        $ \wedge$ $partOf(h,g) $\newline
                        $\wedge$ $studentAt(Tom,g) $\newline
                        $\wedge$ $studentAt(Laura,g) )$,\newline
        where:
        \begin{itemize}
            \item $MB_{i,j}(\phi)$\footnote{See mutual belief's definition in section 3.1} stands for "$\phi$ is a shared belief between agents $i$ and $j$, on $i$'s point of view",
            \item $frontOf(x,y)$ stands for "$x$ is located in front of $y$",
            \item $basketballCourt(x)$ stands for "$x$ is a basketball court",
            \item $partOf(x,y)$ stands for "$x$ is part of $y$",
            \item $studentAt(x,y)$ stands for "$y$ goes or has been at school $y$".
        \end{itemize}
    \item[$\bullet$]  Tom's choice should also be explained by the following weaker belief state:\newline
                        $Bel_{Tom}(MBel_{Laura,Tom} ( frontOf(l,h) $\newline
                        $\wedge$ $basketballCourt(h) $\newline
                        $\wedge$ $partOf(h,g) $\newline
                        $\wedge$ $studentAt(Tom,g) $\newline
                        $\wedge$ $studentAt(Laura,g))$\newline
                        where $B_{i}(p)$ stands for "$i$  believes (that) $p$".
\end{itemize}

\indent The main assumption behind this kind of approach is the rationality and the cooperativeness of dialogue participants.
%infer from ... that ...
In addition, to infer from the fact that someone utters that $p$ that she must also believe that $p$ is commonly assumed as a general rule \cite{Lee97}.
 Nonetheless, this assumption is difficult to handle in practice, as J.A. Taylor et al. have shown \cite{TCM96}, mainly because of
 the computational complexity involved. Furthermore, they proved that, in most cases, nested beliefs are not necessary beyond the second
 level of nesting (ie. what an agent thinks another agent thinks a third agent (possibly the first one) thinks), as long as deception
 is not involved. In the particular case of reference, deception may be involved, as the following situation exemplify, and then may
 require the handling of deeply nested belief.\newline
\newline
\indent Tom and Laura live both in Berlin. They lunched at a restaurant called "Chez Dominique".
    Following this meal, one may reasonably assume that:
    \begin{itemize}
        \item[$\bullet$] $Bel_{Laura}( name(l) = $" Chez Dominique "$ )$,
        \item[$\bullet$] $Bel_{Tom}( name(l) =$ " Chez Dominique "$ )$,
        \item[$\bullet$] And $MBel_{Tom, Laura}( name(l) =$ " Chez Dominique "$.$\newline
        We only treat the particular case of definite reference, which counts as an indication to access a mental
            representation of the intended referent that is supposed to be uniquely identifiable for the hearer. So, it can
            be viewed as a result of a function.
    \end{itemize}

    \indent Then, Laura left Berlin for two years. During this period, the restaurant changed name.  Its new name is "Restaurant la Petite
    Maison".  Tom knows it, but Laura does not know it. Thus, the following situation holds:
    \begin{itemize}
        \item[$\bullet$] $Bel_{Tom}( name(l) =$ " Restaurant la Petite Maison " $)$,
        \item[$\bullet$] $Bel_{Laura}( name(l) =$  " Chez Dominique "$)$.
    \end{itemize}

    The return-day Laura and Tom (who did not leave Berlin) must lunch together. They speak by phone in order to agree upon a time
    and a restaurant. Let's consider the following exchange between them\label{ExampleDialogue}:
    \begin{itemize}
        \item[] $\cdots$
        \item[\textbf{(U1)}] Laura: " Will we lunch at the restaurant where we have been yet ? "
        \item[\textbf{(U2)}] Tom: " Which one ? "
        \item[\textbf{(U3)}] Laura: " \emph{Chez Dominique}. "
        \item[\textbf{(U4)}] Tom: " Ok. "
        \item[] $\cdots$
    \end{itemize}
    \indent At the end of this talk, a conceptual pact of conceptualizing the restaurant as "the place called \emph{Chez Dominique}" is established.
    If we consider that the Conversational Common Ground has to be modelled in terms of mutual belief, the following mutual belief has been
    formed, at least on Laura's point of view: $MBel_{Laura,Tom}(name(l) =$ "Chez Dominique"$)$. Tom's choice of the referring
    expression can not be based on Tom's point of view on the beliefs shared with Laura, because from $MBel_{Tom,Laura}(name(l) =$ "Chez
    Dominique"$)$, one may infer, following mutual belief's definition (ie. \ref{DefMB}) that
    $Bel_{Tom}(name(l) =$ "Chez Dominique "$)$ which is incoherent with $Bel_{Tom}(name(l) =$ "Restaurant la Petite Maison"$)$.
    In fact, Tom's choice should be explained in terms of his nested belief: $Bel_{Tom}(MBel_{Laura,Tom}(name(l) =$ "Chez Dominique"$))$
    and this is a case of deception.\newline
    \newline
    \indent According to previous work \cite{Sag06}, we claim that such a treatment of reference, depending on beliefs of
    dialogue participants at the first place, which may lead to computational representation and treatment with high
    complexity, are neither necessary, nor proper. The proper social attitude is Collective Acceptance.

\subsection{Collective Acceptance, reference and subordinated activity}

\indent Modeling conceptual pacts in terms of belief states implies that the literal description has to be true, or, more precisely,
consistent with dialog partners' beliefs (at least with %(speaker's beliefs on)
shared beliefs between dialog partners on addressee's point of view),
 in order to ensure their rationality. But the goal of Tom and Laura, in our preceding examples, is to determine a
place in such manner that each one identifies it correctly; then, they will be able to meet at the correct meeting-place. Their
goal is not to establish the truth with respect to the place in question. Actually, the establishment of conceptual pacts is governed by the " grounding criterion " \cite{CS89}:\emph{ " The contributor
and the partners mutually believe that the partners have understood what the contributor meant to a criterion sufficient for the
current purpose."} Thereby, one can establish a conceptual pact in conflict with
ones own beliefs, if this pact enables each group member concerned to achieve the current common goal. In the first example, one can imagine
that the basketball-court does not exists any more, and that both Tom and Laura know this fact. Tom's utterance and Laura's agreement
are still realistic.\newline
\indent Collective acceptance of a group of agents, in contrast with belief, may be inconsistent with their beliefs (individual or
shared beliefs). In fact, a description is accepted by the addressee if
it allows him to identify the intented referent and if an inconsistent description is not an obstacle to the realization of current
goals. For example, if Tom have to send a letter to Laura (having two postal addresses) and say: "Must I send you the letter at 16
Collingham Road, London". Even if Laura identifies the correct place, the address has to be correct to receive Tom's letter. Finally,
conceptual pact is a temporary and flexible concept, this property does not match with the ideal of integration or agglomeration of
beliefs.\newline
\newline
\indent How explaining the use of Collective Acceptance ? Generally, this may be partly due to a particular aspect of
goal-oriented dialog as a subordinated sub-activity. Goal-oriented dialogues are implied by two interdependent
collaborative activities, as explained by A. Bangerter et al.: \emph{"Dialogues, therefore, divide into two
planes of activity \cite{Cla96}. On one plane, people create dialogue \textbf{in service of} the basic joint activities they are engaged
in-making dinner, dealing with the emergency, operating the ship. On a second plane, they manage the dialogue itself-deciding who
speaks when, establishing that an utterance has been understood, etc. These two planes are not independent, for problems in the
dialogue may have their source in the joint activity the dialogue is in service of, and vice versa. Still, in this view, basic joint
activities are primary, and dialogue is created to manage them."}\footnote{This claim must be extended to other kind of
basic activity such as cooperative or competitive activities.} \cite{BC03}. One of team members'goals is to understand each
other, in other words to reach a certain degree of intelligibility, sufficient for the current purpose. \newline
\newline
\indent One may distinguish between two kinds of sub-activities: sub-activities which are sub-parts of another activity (thus, which
transcripts the compositionality of basic activities) and sub-activities \emph{in service of} another activity, ie. subordinated (sub-)activities,
such as planning, problem solving, interaction with other agents (goal-oriented dialog) and so on. On the logical point of view, the rationality
of the involved agents is rendered by a coherent mental state and by the notion of rational action \cite{CL90,Sad94b}. For example, the
beliefs and intentions of an agent form a consistent set and agent's actions are also consistent with his beliefs and intentions. At the first glance,
the coherence of action with beliefs seems to be irrefutable. However, to the extend that the success of a subordinated activity is
governed by the generalization of the sufficient criterion and on the basis of preceding arguments, one may reasonably assume that
agents' rationality does not strictly imply the coherence between the actions being parts of a subordinated activity and the beliefs states of
the involved agents. For these reasons, conceptual pact match better with acceptance and modeling conceptual pacts by collective
acceptance insure the rationality of team members. However, agent's rationality is contingent on the motivational context and on the context of mental states
of dialog partners.

    \subsection{The philosophical notion of Collective Acceptance}

\indent Studies on dialog modeling as a collaborative activity address the philosophical problem of determining the type of mental
states which could be ascribed to team members. Based on the observation that sometimes one may encounter situations
where one has to make judgements or has to produce utterances that are contrary to ones privately held beliefs, philosophers, such has
\cite{Coh92}, have introduced the notion of (Collective) Acceptance, which is an intentional social mental attitude.
(Collective) Acceptances have the following properties, in contrast with beliefs \cite{Wra01}:
    \begin{itemize}
        \item[$\bullet$] They are voluntary (or intentional);
        \item[$\bullet$] They holds on utility or success (thus we can accept something we believe false);
        \item[$\bullet$] They does not required justifications;
        \item[$\bullet$] All or nothing: we decide to accept or not to accept.
    \end{itemize}

In J.L. Cohen's famous book, "An essay on belief and acceptance" \cite{Coh92}, the author argue that the conversational implicature
"a person's saying that $p$ implies that this person believes $p$" is not the rule and that speech acts such as concessions,
acknowledgements, agreements and admissions that $p$ do not imply the existence of the corresponding belief. In such cases, "I thereby accept
that $p$" means that "I take that proposition as a premiss for any relevant decision or argument" \cite{Coh92}. In previous work \cite{Sag06},
we claim that an act of reference using a particular description $\imath x.descr(x)$ of an object $o$ does not imply that the speaker
believes that $\imath o.descr(o)$ holds, but implies that the speaker believes that this description enables the addressee to pick out
the correct intented referent.

\section{Formal part}

    \subsection{The dialog model}

Rational models, based on \cite{CL90}, can be considered as a logical reformulation of plan-based models. They integrate, in more, a
precise formalization of dialog partners' mental states (their beliefs, choices (or desires) and intentions), of the rational balance
which relates mental attitudes between them and relates mental attitudes with agents' acts. Moreover, dialogue acts' preconditions
and effects are expressed in terms of dialog partners' mental states. Thus, this is hopeful to model precisely mental attitudes.
\newline
\newline
\indent The chosen model is based on the rational model proposed by D. Sadek \cite{Sad94b}, extending \cite{CL90}, which rests upon a set of principles
(axiom schemas) of which dialog acts are branched off.
A dialog system is considered as a cognitive agent which is rational and have a cooperative
attitude towards other agents (as the dialog system's users) and this agent is able to communicate with other agents.\newline
\indent Mental states (beliefs, intentions,...) and actions are formalized in a first-order modal logic. In the following of the paper,
the symbols $\neg, \wedge , \vee, \Rightarrow$ stand for the connectors of the classical logic (respectively negation, conjunction,
disjunction and implication); $\forall, \exists$ stand for the universal  and existential quantificators; $p$ stands for a closed formula
denoting a proposition; $i, j$  denote agents and $\phi$ is a formula schemata. We only need to introduce
here two mental attitudes, belief and intention:\newline
\newline
\indent $B_{i}(p)$ stands for "$i$ (implicitly) believes (that) $p$",\newline
\indent $I_{i}(p)$ stands for "$i$ intends to bring about $p$".\newline
\newline
\indent
Action expressions can be formed with primitive acts: with $(a_{1}; a_{2})$ which stands for sequential action (where $a_{1}$ and $a_{2}$ are action
expressions) and with $(a_{1} | a_{2})$ which stands for non-deterministic choice.
    \begin{itemize}
            % non utilisé par la suite
            %\item[$\bullet$] $Feasible(a,p)$: "$a$ can take place, and, and, if it does, then $p$ will be subsequently be true"\newline
            %                $Feasible(a) = Feasible(a,true)$ \newline
            \item[] $Done(a,p)$: "$a$ has just taken place, and $p$ was true before that" \newline
                            $Done(a) = Done(a,true)$
    \end{itemize}
\indent The model of communicative acts is:
\begin{itemize}
    \item[] $<i,TypeOfCommunicativeAct(j,\phi)>$
    \item[] FP: "Feasible Preconditions": the conditions which must be satisfied in order to plan the act;
    \item[] PE: "Perlocutionary Effect": the reason for which the act is selected.
\end{itemize}

For example, the communicative model of "$i$ informing $j$ that $p$" is:
\begin{itemize}
    \item[] $<i,INFORM(j,\phi)>$
    \item[] FP: $B_{i}(\phi) \wedge \neg B_{i}(B_{j}(\phi))$
    \item[] PE: $B_{j}(\phi)$
\end{itemize}

\indent In this model, utterance generation and understanding, and thus referential acts are considered as individual acts. Furthermore,
the perlocutionary effects are considered as achieved as soon as the communicative act has been performed.\newline
\indent So dialog and reference treatment are not considered as collaborative activities. In order to do so, notably, the set of mental
attitudes has to be extended with notions such as collective intention and mutual belief.
\newline
\indent There is no consensus on the definition of collaboration. We consider that a group of agents is engaged in a collaborative
activity as soon as they share a collective intention.\newline
\newline
\indent $CollInt_{i,j}(\phi)$ stands for "$i$ and $j$ collectively intends to bring about $p$, on $i$'s point of view".\newline
\newline
\indent $MB_{i,j}(\phi)$ stands for "$\phi$ is a shared belief between agents $i$ and $j$, on $i$'s point of view" and
mutual beliefs are formalized as\label{DefMB}:

 \begin{center}
 $MB_{i,j}(\phi)\equiv Bel_{i}(\phi \wedge MBel_{j,i}(\phi))$\end{center}

Furthermore, Collective Acceptance have to be included.

    \subsection{Collective Acceptance}

    We propose the following formalization of the philosophical notion of Collective Acceptance:
\begin{itemize}
    \item[$\bullet$]  $CollAcc_{ij}(\phi)$ stands for "$\phi$ is a collective acceptation between agents $i$ and $j$, on $i$'s point of
    view"
    \item[$\bullet$]  Collective Acceptance is an intentional attitude, ie. it comes from individual acts of involved agents:\newline
        $((\exists \alpha,\beta \in \{i,j\}).$\newline
        $Done(Prop_{\alpha \beta}(\phi)) \wedge Done(Accept_{\beta \alpha}(\phi)) )$\newline
        $\Rightarrow CollAcc_{ij}(\phi)$\newline
        where:\begin{itemize}
            \item $Prop_{ij}(\phi)$ stands for "$i$ proposes $j$ to consider $\phi$"
            \item $Accept_{ji}(\phi)$ stands for "$j$ accepts to consider $\phi$ (towards $i$)"
            \item $Prop_{ij}(\phi)$ and $Accept_{ji}(\phi)$ are individual actions.
        \end{itemize}
    \item[$\bullet$]  A proposition involves a social obligation to react:\newline
        $Done(Prop_{i,j}(\phi))$\newline
        $ \Rightarrow (I_{j}(Done(($
            \begin{itemize}
                \item[] $\hspace{1cm}$ $Accept_{j,i}(\phi)$
                \item[] $\hspace{1cm}$ $| (Prop_{j,i}(\phi'))$
                \item[] $\hspace{1cm}$ $| (request_{j,i}(Prop_{i,j}(\phi'')))))$
            \end{itemize}
     $\hspace{0.5cm}$ $\wedge ((\phi' \neq \phi) \wedge (\phi'' \neq \phi)))$
\end{itemize}
\indent Following \cite{BDL00}, we consider that social obligations as pro-attitudes are not required and that an anticipatory coordination
takes place on the speaker's point of view. This phenomenon is govern by a social rule, acquired during preceding social interaction.
This social rule is transcribed by repeated use through a reaction to the realization of a particular action (on the speaker's point of
view) and through a reaction to the observation of an event which is the occurrence of a particular action (on the addressee's point
of view). Since, reaction is a unintentional action, we have to extend the kind of action of the basic model. In fact, this
model only considers what we name \emph{intentional actions}. Intentional actions of an agent are those generated by a chain of
intention, in our model they are generated by the activation of the rational axiom \cite{Sad94b}:
\begin{itemize}
    \item[] $I_{i}(p) \Rightarrow I_{i}(Done(a_{1}\vee \cdots\vee a_{n}))$ \newline
                            The intention of an agent, to achieve a given goal, generates the intention that one of the acts,
                            which satisfies the following conditions, be performed:
                            \begin{enumerate}
                                \item $(\exists x)B_{i}(a_{k} = x)$ $\equiv$ $Bref_{i}(a_{k})$: \\
                                    the agent $i$ knows the action $a_{k}$,
                                \item $EP_{a_{k}}=p$ and
                                \item $\neg I_{i}(\neg Possible(Done(a_{k})))$
                            \end{enumerate}
\end{itemize}

\indent
\emph{Reactions} have to be added. Reactions of an agent are defined as those generated by the activation of such axiom:
\begin{itemize}
    \item[] $\phi \Rightarrow I_{i}(Done(a_{1}\vee \cdots\vee a_{n}))$ \newline
        where $\phi$ is the result of the perception of an event or an action's occurrence.
\end{itemize}

\section{Model of Reference as a collaborative activity}

    \subsection{Model of Referential Act}

\indent In order to model dialog as a collaboration, reference treatment has to be considered at the speech act level \cite{CL94}, as
it is done in A. Kronfeld's work \cite{Kro90}.\footnote{For a computational implementation is provided in \cite{Jor00}.}\newline
\indent In order to integrate Collective Acceptance in reference, we propose an extension of an existing model of referential acts
        based on A. Kronfeld's work in the rational model used \cite{BPS95}. The act of reference from an agent $i$ to another agent
        $j$, using the conceptualization $x$ (which corresponds to the semantics of the referential expression) to refer to the object $y$ is formalized as:
        \begin{itemize}
            \item[] $<i, REFER(j,x,o)>$
            \item[] FP: $I_{i}(refer_{i,j}(o)) \wedge Bref_{i}(o)$;
            \item[] EP: $B_{j}((\exists o) I_{i}(refer_{i,j}(o))) $
                            \begin{itemize}
                                \item[] $\wedge$ $I_{j}(Bref_{j}(o)$)
                                \item[] $\wedge$ $RepSameObj(o,o')$
                                \item[] $\wedge$ $Done(Prop_{i,j}(referBy(x,o)))$
                                \item[] $\wedge B_{j}(Done(Prop_{i,j}(referBy(x,o))))$
                            \end{itemize}
            where:
            \begin{itemize}
                \item $o$ et $o'$  are object mental representations;
                \item $I_{i}(refer_{i,j}(o))$ stands for "a communicative intention of $i$ to refer to $o$, the addressee is $j$";
                \item $RepSameObj(o,o')$ stands for "the mental representations $o$ and $o'$ represent the same object";
                \item $referedBy(D,R)$ stands for "the description $D$ refers to the referent $R$".
            \end{itemize}
        \end{itemize}
        %$PC(x,o)$ stands for "the conceptual pacts: conceptualizing $o$ as the only object such as $x$".
        \indent Generating a referential expression is considered as the generation of an instance of such plan and the interpretation of a
        particular referential expression as the recognition of an instance of such plan. And the whole process is governed by two
        meta-goal, on the speaker's point of view \footnote{On the addressee point of view, it is govern by dual goals with existential
        quantifier.}:
                \begin{center}
                $CollInt_{ij}(MB_{ij}(I_{i}(refer_{i,j}(o)))) \wedge CollInt_{ij}((\exists D)CollAcc_{ij}(referedBy(D,o)))$\end{center}

    \subsection{Return to the example}

Let's consider the example shown in \ref{ExampleDialogue}, the task level and the conversational level have to be separated
\footnote{Further details may be found in\cite{Sag06}.}.
In uttering (U1), Tom wants to make a necessary choice for the meeting task, such as :
\begin{center}
    $I_{Tom}((\exists l)MBel_{Tom, Laura} ( meetingPlace = l ) ) $    \newline
\end{center}
\indent Tom makes his choice: his mental representation of the restaurant chosen is $o$. In order to realize his preceding intention,
he has get through to Laura:
\begin{center}
$I_{Tom}(refer_{Tom,Laura}(l))$  \newline
\end{center}
\indent Remaining the goal of referential acts (\ref{GoalOfRefer}), the choice of the description of the intented place is guided by
its capacity to enable Laura to pick out, \emph{in her mental state}, the mental representation of the correct place. That is,
the description enables Laura to isolate the correct mental representation from other possible ones, with sufficient evidence of mutuality.
This is a pragmatic (ie. contextual) guideline, which corresponds to the Identification goal. \newline
\newline
\indent Thus, Tom produces a description of the intended place: \emph{"the restaurant where we have been yet"}. He thinks that
Laura is able to identify the correct place basing on the description, ie. he thinks that she is able to realize the following
intention:
\begin{center}
Identification task: $I_{Laura}(Bref_{Laura}(l')\wedge RepSameObj(o,o'))$
\end{center}
But, Laura is not able to pick out a single place: there is other restaurants, where they have been together. Moreover, Laura
has to answer to Tom's proposition:
\begin{center}
$B_{Laura}(Done(Prop_{Tom,Laura}(referedBy(\imath x. \phi(x),l'))).$   \newline
\end{center}
\indent She is obliged to reply to his proposition by the social rule. Besides, the precondition of accepting a conceptual pact is to
have realized the Identification goal; otherwise, the addressee has the choice between other possible reactions. As Laura
failed to succeed, she chooses to ask for clarification in (U2):
\begin{center}
$request_{Laura,Tom}$\newline
$(Prop_{Tom,Laura}(referedBy(\imath x. \phi'(x),l'))).$ \newline
$\wedge (\phi ' \neq \phi)$
\end{center}
\indent In order to achieve understanding, by a cooperative attitude, Tom realizes Laura's request in (U3). Laura is now able to
pick out a single mental representation of the place. She likes it, so she agrees. The social goal obliges Laura to react to
Tom's new proposition. As the precondition of accepting is fulfilled, with uttering (U4), Laura realizes the following intention:\begin{center}
$Done(Accept_{Laura,Tom}(referedBy(\imath x. \phi'(x),l'))).$
\end{center}
Finally, following Collective Acceptance's definition, a conceptual pact is created:
\begin{center}
$CollAcc_{Laura,Tom}(referedBy(\imath x. \phi'(x),l').$\end{center}
As well as, mutual understanding:
\begin{center}
$MB_{Laura,Tom}(I_{Tom}(refer_{Tom,Laura}(l')),$
\end{center}
and the coordination on the task level:
\begin{center}
$MBel_{Laura,Tom} ( meetingPlace = l' ) ) $.  \end{center}

\section{Conclusion}

\indent Modeling dialog as a collaborative activity consists notably in specifying the content of the Conversational Common Ground and the kind of
social mental state involved. Even if mutual beliefs, or weaker forms of belief states, do not rise to inconsistencies, but, are still sufficiently strong for
the participants to have successful cooperation or coordination of actions. Epistemic states involve computational treatments with high complexity.\newline
\indent We show that modeling the CCG by an epistemic state is neither necessary, nor proper. Considering only genuine conceptual pacts limits
the capacity of interaction and may leads to "real" communicative errors.\newline
\indent We have proposed a formalization of Collective Acceptance, furthermore, elements haven been given in order to
integrate this attitude in a rational model of dialog. Finally, a model of referential acts as being part of a
collaborative activity has been provided.\newline
\newline
\indent Further studies will hold on the extension of the general principles proposed to the dialog itself. Moreover, collective acceptance is a
particularly interesting attitude because it allows to model reference and dialog itself as situated activities in an elegant manner.
Finally, this concept may provide symbolic elements in order to form the grounding criterion, which is a notion especially hard to make up,
 because this criterion is highly context dependant. Grounding criterion differs depending on the people involved, the domain concerned and so on.
\newline
\newline
\textbf{Acknowledgment} \newline
This work is partially financed by the grant A3CB22 / 2004 96 70 of the regional council of Brittany.

\bibliographystyle{acl}
\bibliography{bibArticleSagetBrandial}

\end{document}